%% file: main.tex
\documentclass{article}

\PassOptionsToPackage{numbers, compress}{natbib}
\usepackage{graphicx}
\usepackage{amsmath}
\usepackage{amssymb}
\usepackage{booktabs}
\usepackage{currfile}
\usepackage{xcolor}
\usepackage{multirow, enumitem, subcaption, float, wrapfig}
\definecolor{forestgreen}{rgb}{0.13, 0.55, 0.13}

\usepackage[pagebackref,breaklinks,colorlinks]{hyperref}
\usepackage[capitalize]{cleveref}
\crefname{section}{Sec.}{Secs.}
\Crefname{section}{Section}{Sections}
\Crefname{table}{Table}{Tables}
\crefname{table}{Tab.}{Tabs.}

\usepackage[preprint]{neurips_2024}

\usepackage[utf8]{inputenc} % allow utf-8 input
\usepackage[T1]{fontenc}    % use 8-bit T1 fonts
\usepackage{hyperref}       % hyperlinks
\usepackage{url}            % simple URL typesetting
\usepackage{booktabs}       % professional-quality tables
\usepackage{amsfonts}       % blackboard math symbols
\usepackage{nicefrac}       % compact symbols for 1/2, etc.
\usepackage{microtype}      % microtypography
\usepackage{xcolor}         % colors

\title{VidLPRO: A \underline{Vid}eo-\underline{L}anguage \underline{P}re-training Framework for \underline{Ro}botic and Laparoscopic Surgery}

\author{%
Mohammadmahdi Honarmand\thanks{Work done while Mahdi was an Intern at Intuitive Surgical Inc.}\\
Stanford University
\And
Muhammad Abdullah Jamal, Omid Mohareri\thanks{Corresponding author}\\
Intuitive Surgical Inc.\\
}%

\begin{document}

\maketitle

\begin{abstract}

We introduce VidLPRO, a novel video-language (VL) pre-training framework designed specifically for robotic and laparoscopic surgery. While existing surgical VL models primarily rely on contrastive learning, we propose a more comprehensive approach to capture the intricate temporal dynamics and align video with language. VidLPRO integrates video-text contrastive learning, video-text matching, and masked language modeling objectives to learn rich VL representations. To support this framework, we present GenSurg+, a carefully curated dataset derived from GenSurgery, comprising 17k surgical video clips paired with captions generated by GPT-4 using transcripts extracted by the Whisper model. This dataset addresses the need for large-scale, high-quality VL data in the surgical domain. Extensive experiments on benchmark datasets, including Cholec80 and AutoLaparo, demonstrate the efficacy of our approach. VidLPRO achieves state-of-the-art performance in zero-shot surgical phase recognition, significantly outperforming existing surgical VL models such as SurgVLP and HecVL. Our model demonstrates improvements of up to 21.5\% in accuracy and 15.7\% in F1 score, setting a new benchmark in the field. Notably, VidLPRO exhibits robust performance even with single-frame inference, while effectively scaling with increased temporal context. Ablation studies reveal the impact of frame sampling strategies on model performance and computational efficiency. These results underscore VidLPRO's potential as a foundation model for surgical video understanding.

\end{abstract}

\section{Introduction}
\label{sec:intro}

%\subsection{Importance of AI in Surgical Computer Vision} 
The field of surgical computer vision has seen significant advancements in recent years, driven by the growing demand for artificial intelligence (AI) applications in healthcare. A notable increase in research has led to the development of deep learning models that enable surgical workflow recognition~\cite{Blum2008,Blum2010,2016_Dergachyova}, enhance surgical scene understanding~\cite{Nwoye_scene,alapatt2021,allan2019} and reconstruction~\cite{rivoir2021,wang2022,Pfeiffer2019}. As surgical procedures grow more complex and technology-driven, the demand for intelligent systems that support surgeons throughout the entire surgical journey - from preoperative planning to intra-operative guidance and post-operative analysis~\cite{padoy2019machine} - becomes increasingly crucial for enhancing patient outcomes, streamlining workflows, and enhance overall surgical efficiency~\cite{maier2017surgical,vercauteren2019cai4cai}.

Despite these promising applications, the development and implementation of these systems in surgical domain face several challenges. One of the primary challenges is the complexity and variability inherent in surgical procedures. Unlike many standardized video datasets, surgical videos capture highly dynamic environments where the visual content can vary significantly based on the specific procedure, patient anatomy, surgeon technique, and unexpected complications \cite{jin2017sv,twinanda2016endonet}. This variability makes it difficult to develop robust models that can generalize across different surgical scenarios. Another significant challenge is the scarcity of large-scale annotated surgical datasets. Unlike in other domains where data can be more readily collected and labeled, surgical data is subject to strict privacy regulations and requires expert annotation, which is both time-consuming and expensive \cite{maier2017surgical,bodenstedt2018comparative}. This limitation hinders the development of data-hungry deep learning models and necessitates innovative approaches to leverage limited labeled data effectively. The long duration of surgical procedures also poses a unique challenge. Surgical videos often span several hours, requiring models to capture and process long-range temporal dependencies \cite{padoy2019machine}. This is in stark contrast to many general video understanding tasks that typically deal with short clips lasting only a few seconds or minutes. Furthermore, interpreting surgical videos requires specialized medical knowledge, making it challenging to apply general-purpose video understanding models directly to surgical tasks \cite{vercauteren2019cai4cai}.Lastly, the fine-grained nature of surgical actions and the subtle visual cues that distinguish different phases or steps of a procedure add another layer of complexity. Models must be capable of detecting and interpreting small but crucial details in the surgical field, often in the presence of occlusions, reflections, and rapid camera movements \cite{gao2014jhu,zia2021surgical}. 

% The need for domain-specific expertise extends to the annotation process as well, further complicating the creation of large-scale datasets. Lastly, the fine-grained nature of surgical actions and the subtle visual cues that distinguish different phases or steps of a procedure add another layer of complexity. Models must be capable of detecting and interpreting small but crucial details in the surgical field, often in the presence of occlusions, reflections, and rapid camera movements \cite{gao2014jhu,zia2021surgical}.% To address these challenges, researchers have explored various approaches in surgical video analysis. Traditional methods often relied on supervised learning with hand-crafted features, requiring extensive manual annotation and domain expertise. However, these approaches still face limitations in terms of generalization across different procedures and the need for large amounts of labeled data.

Recently, Multimodal learning, which integrates multiple modalities such visual data, text data, audio, depth maps etc., has emerged as a viable strategy in computer vision domain. Specifically, Vision-Language Pre-training (VLP) which leverages large-scale datasets of paired visual and free-from textual data, can reduce the reliance on annotated datasets, enabling more efficient and effective learning. It enables models to learn rich and generalizable representations that can be adapted to various downstream tasks with minimal fine-tuning such as image-text retrieval~\cite{wu2023multimodal,blip2,Sigmoidloss}, visual question answering~\cite{Yu_2019_CVPR,Singh_2019_CVPR,VisualBERT,ViLBERT,UNITER}, video understanding~\cite{wang2023internvid,VideoPrism,li2023lavender,fu2021violet,MERLOT} and zero-shot classification~\cite{radford2021learning,align}. The potential of VLP to capture complex relationships between visual content and natural language descriptions makes it particularly appealing for the surgical domain, where procedures are often accompanied by detailed textual reports or narrations.

\begin{wrapfigure}{r}{0.5\textwidth}
  \vspace{-20pt}
  \setlength{\abovecaptionskip}{-1pt}
  \setlength{\belowcaptionskip}{-0.3cm}  
  %\vspace{-40pt}
  \centering
  \includegraphics[width=0.45\textwidth]{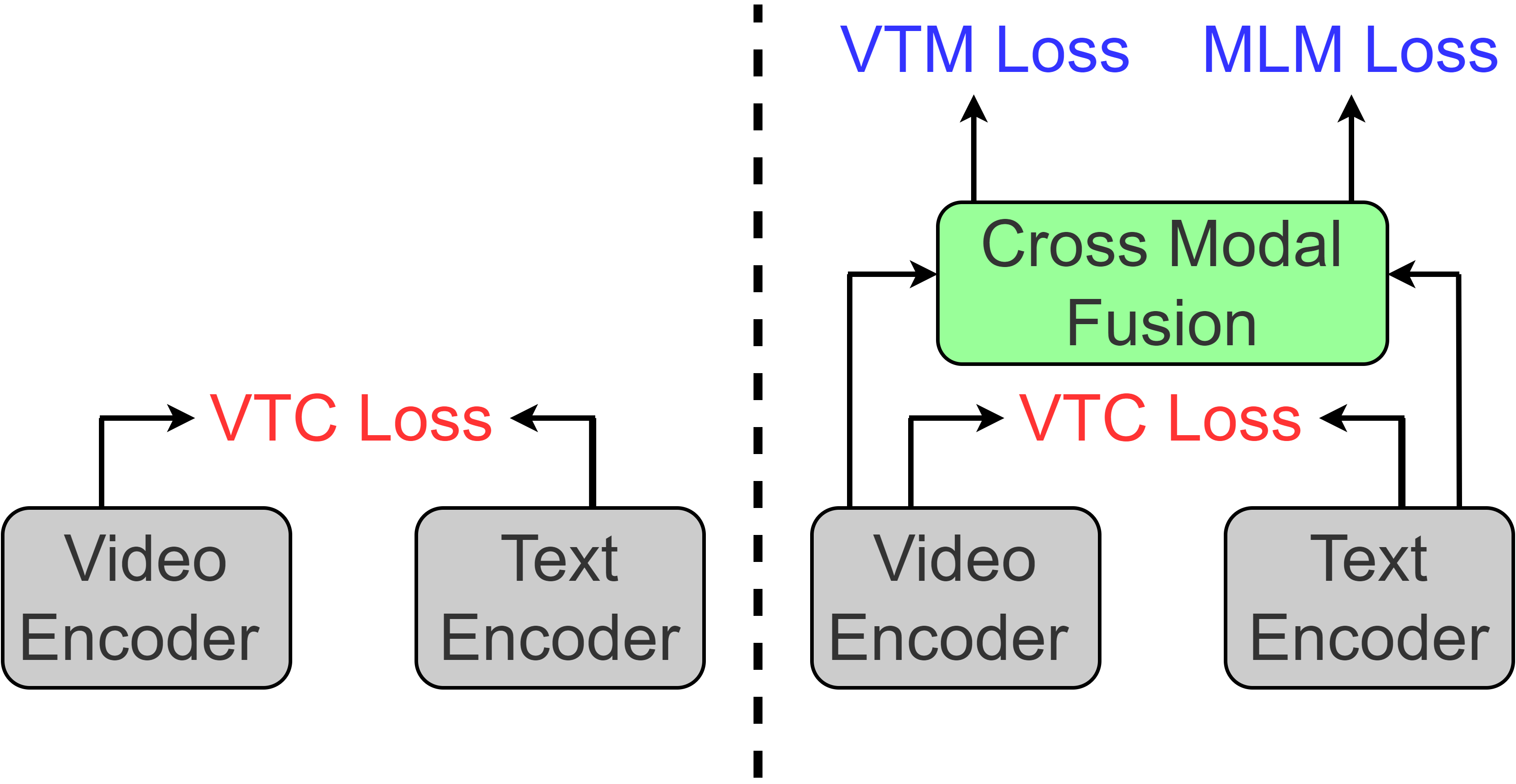}
  % \vspace{0pt}
  \caption{\small Current approaches (\textbf{left}) rely on video-text contrastive loss only, while our method (\textbf{right}), besides contrastive loss, employ video-text matching loss and masked language modeling to enhance cross-modal fusion and surgical language.}
  %\vspace{-15pt}
  \label{fig:comparison}
\end{wrapfigure}

Recent efforts have begun to explore the application of VLP techniques to surgical video analysis. Notable approaches include SurgVLP \cite{yuan2023learning}, which leverages surgical video lectures and their transcripts to learn multi-modal representations, and HecVL \cite{yuan2024hecvl}, which proposes a hierarchical pre-training framework for zero-shot surgical phase recognition. While these methods have shown promising results, they still face several limitations. A significant challenge has been the lack of large-scale, diverse datasets for surgical VLP. The introduction of the GenSurgery dataset \cite{schmidgall2024general} was a step forward, providing a substantial collection of surgical videos. However, this dataset had limitations, including a lack of paired textual data, inconsistent audio quality, and the presence of non-informative content. Our GenSurg+ dataset addresses these issues by rigorously filtering the original data, adding high-quality captions, and ensuring rich linguistic context. Despite this progress, existing approaches still struggle with insufficient temporal modeling, failing to capture long-range dependencies in surgical videos effectively. Many current methods show reduced performance when applied to new surgical procedures or tasks not seen during pre-training, indicating limited generalization capabilities. Additionally, most approaches rely solely on video-text contrastive (VTC) learning as shown in Figure~\ref{fig:comparison}, missing out on the benefits of other pretraining objectives that could enhance the model's understanding of surgical content and context. Addressing these limitations is crucial for advancing the field of surgical VLP and developing more robust and versatile models for surgical video understanding.

%\AJ{Maybe write the limitations in the paragraph. Try to make it concise but overall story should be conveyed.}

%\AJ{Also talk about original GenSurgey dataset first where you mention lack of large-scale dataset. }

To address the limitations of existing surgical VLP approaches, we present VidLPRO and GenSurg+, a novel framework and dataset for robotic and laparoscopic surgical video-language foundation models. VidLPRO builds upon recent advancements in video-language pre-training, incorporating a Vision Transformer (ViT) as the video encoder, BERT as the text encoder, and a multimodal fusion module. Our model employs a combination of Video-Text Contrastive Learning (VTC), Video-Text Matching (VTM), and Masked Language Modeling (MLM) objectives to learn nuanced, context-aware representations of surgical procedures as shown in Figure~\ref{fig:comparison}. We also introduce GenSurg+, an enhanced version of the GenSurgery dataset~\cite{schmidgall2024general}, containing \textbf{17k} 45-second clips of endoscopic robotic surgery with high quality captions generated using raw narration and GPT-4. In zero-shot surgical phase recognition, VidLPRO significantly outperforms the current state-of-the-art on both Cholec80~\cite{twinanda2016endonet} and AutoLaparo~\cite{wang2022autolaparo}. More specifically, on Cholec80, it achieves \textbf{57.1\%} accuracy and \textbf{32.1\%} F1 score, surpassing HecVL by \textbf{15.4\%} and \textbf{5.8\%} respectively. Our ablation studies demonstrate VidLPRO's robustness across different frame sampling rates, with performance scaling effectively as frame count increases.
These results highlight the effectiveness of our pre-training approach, the quality of GenSurg+, and VidLPRO's potential to generalize across different surgical procedures and tasks, crucial for developing adaptive AI systems for diverse surgical environments.

\section{Related Work} 

\subsection{Vision-Language models} Most of the vision-language approaches can be categorized into two groups. The first group focuses on training multi-modal encoders~\cite{SOHO,VinVL,UnicoderVL,ImageBERT,VLBERT} while second group focuses on training uni-modal vision and text encoders~\cite{radford2021learning,align,Flava,Florence}. In context of surgical domain, Surgical-VQA~\cite{surgvqa}, SurgicalGPT~\cite{SurgicalGPT} propose vision-language model for visual question answering. Surgical-LVLM~\cite{SurgicalLVLM} adapts large vision-language model by introducing specialized Visual Perception LoRA blocks for grounded visual question answering in robotic surgery.

\subsection{Surgical Video-Language Pretraining} The application of video-language pre-training (VLP) techniques to the surgical domain is a recent development that shows great promise for advancing surgical video analysis. Two notable approaches in this emerging field are SurgVLP \cite{yuan2023learning} and HecVL \cite{yuan2024hecvl}, which have made significant strides in adapting VLP methods to the unique challenges of surgical data. SurgVLP \cite{yuan2023learning} uses contrastive learning objective to learn multi-modal representations from surgical video lectures. This method leverages a large dataset of surgical videos paired with transcribed audio, using multiple complementary automatic speech recognition (ASR) systems to generate text annotations. Building upon this foundation, HecVL \cite{yuan2024hecvl} proposes a hierarchical video-language pre-training framework specifically designed for zero-shot surgical phase recognition. This approach addresses the challenge of capturing both fine-grained actions and high-level surgical concepts by incorporating hierarchical textual supervision. VidLPRO, on the other-hand, introduces multiple pre-training objective beyond mere contrastive learning to capture more rich multi-modal representations.

\subsection{Surgical Phase Recognition} 
Surgical phase recognition aims to automatically identify and segment different stages of a surgical procedure. Traditional approaches to surgical phase recognition often relied on hand-crafted features and classical machine learning techniques~\cite{padoy2012statistical,Blum2010}. However, with the advent of deep learning, there has been a shift towards more sophisticated models that can automatically learn relevant features from raw video data \cite{jin2017sv,Dergachyova2016}. Following these, many one-stage approaches~\cite{SVRCNet,MTRCN,TMRNET} have been proposed to learn spatio-temporal features. However, one-stage approaches fail to capture the long-term spatial-temporal dependency. To address this limitation, two-stage solutions~\cite{czempiel2020tecno,TransSVNet,lovit,SKiT} are proposed which first extract the spatial or spatio-temporal features using the feature extractor and then employ a temporal model on the top of these features to learn long-term dependency. The temporal models are typically categorized into three types: Recurrent Neural Networks (RNNs)~\cite{jin2020multi}, Temporal Convolution Networks (TCNs)~\cite{czempiel2020tecno,Ramesh_2021}, and Transformers~\cite{OperA}.

\section{Method}

\subsection{GenSurg+}
\label{sec:formatting}

To enable effective video-language pre-training for robotic and laparoscopic surgery, we introduce GenSurg+, a large-scale dataset of surgical videos paired with descriptive captions. GenSurg+ builds upon the GenSurgery dataset \cite{schmidgall2024general}, which was originally introduced as the largest publicly available dataset of general surgery videos.

\subsection{Dataset Creation Pipeline} 
\begin{figure*}[t]
\centering
\includegraphics[width=0.95\textwidth]{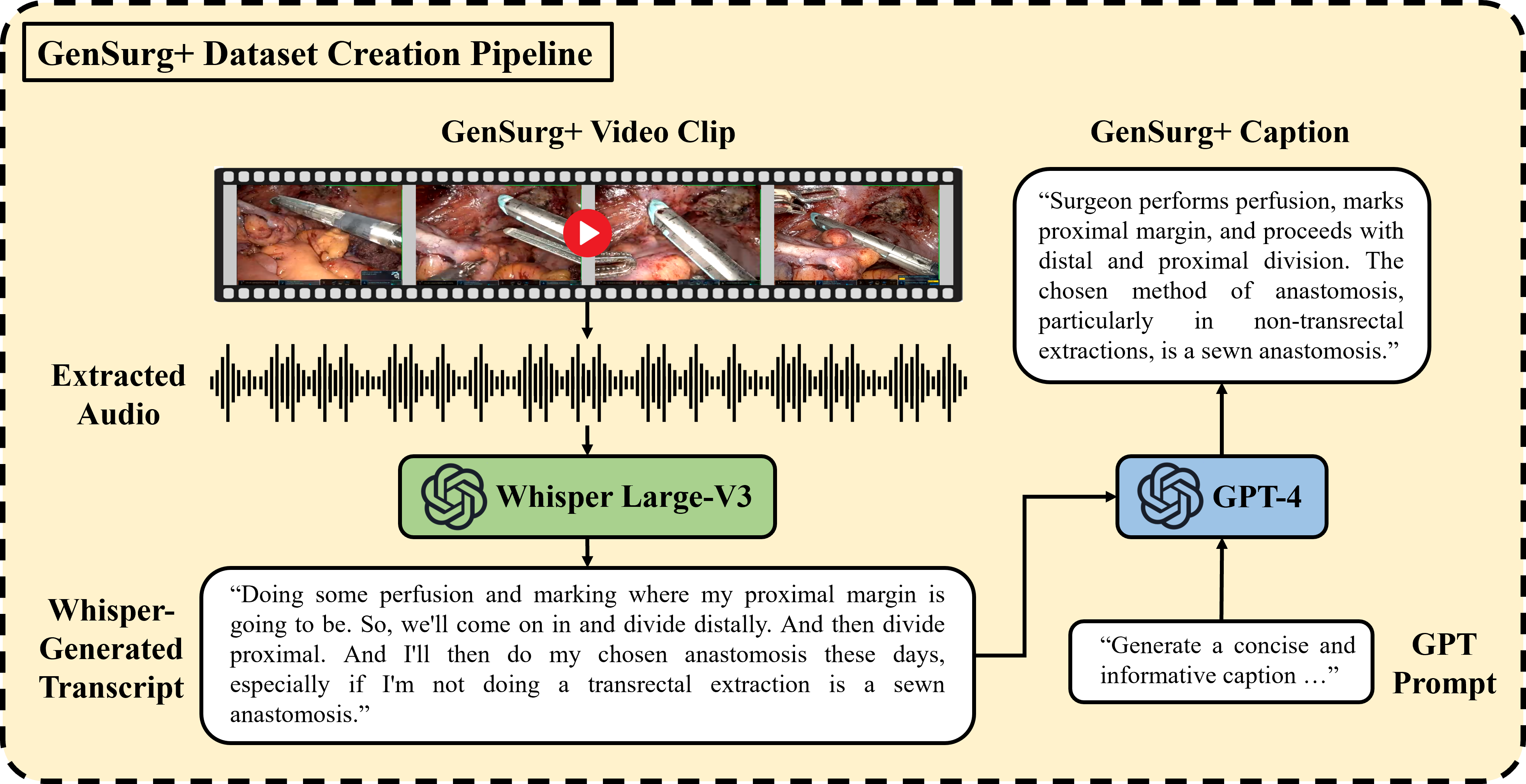}
\caption{Overview of the \textbf{GenSurg+} dataset creation pipeline.}
\label{fig:gensurg_pipeline}
\end{figure*}

We began with the original GenSurgery dataset, which contains 3,100 videos spanning 28 different surgical procedures and totaling 680 hours of content. Our dataset creation pipeline involved several key steps to refine and augment this initial corpus:

%\begin{enumerate}
\paragraph{Audio Filtering.} We first filtered out 1,300 videos that lacked audio content, as audio is crucial for generating meaningful textual descriptions.
\paragraph{Transcript Extraction.} For the remaining 1,800 videos with audio, we employed the Whisper model \cite{radford2022whisper} to extract speech transcripts. This step was necessary as many of the videos, due to their age, lacked reliable YouTube automatic captions.
\paragraph{Video Segmentation and Filtering.} We segmented the videos into 45-second clips, resulting in approximately 18,000 individual segments. To ensure the quality and relevance of our dataset, we further filtered these clips based on linguistic criteria. Specifically, we removed about 1,000 clips that contained either too few unique words or highly repetitive content. This step helped eliminate silent segments and portions with non-informative audio (e.g., background music or noise).
\paragraph{Caption Generation.} For the remaining 17,000 high-quality video clips, we generated descriptive captions using the GPT-4 language model \cite{openai_gpt4}. We crafted a specialized prompt to ensure the captions were concise, informative, and tailored to the surgical domain. Please see appendix for the prompt. The complete pipeline for creating GenSurg+ is illustrated in Figure \ref{fig:gensurg_pipeline}.

\begin{table}[h!]
  \centering
  \caption{Comparison between GenSurg+ and SVL-Pretrain datasets.}
  \begin{tabular}{p{3cm} | p{1.4cm} | p{2.55cm}}
    \toprule
    & GenSurg+ & SVL-Pretrain \cite{yuan2023learning} \\ 
    \midrule
    Publicly Available & $\checkmark$ & $\times$ \\ 
    $\#$ Videos & 1.3k & 1.3k \\ 
    $\#$ Clip-Caption Pairs & 17k & - \\ 
    Total Duration & 213 hours & - \\ 
    \bottomrule
  \end{tabular}
  \label{table:comparison}
\end{table}

\subsection{Dataset Statistics and Characteristics} 

The resulting GenSurg+ dataset comprises 17,000 45-second video clips, totaling 213 hours of high-quality surgical content paired with descriptive captions. As shown in Table \ref{table:comparison}, this makes GenSurg+ the largest publicly available dataset specifically designed for surgical video-language pre-training, offering a significant resource for research in this area.

GenSurg+ represents a significant step forward in enabling large-scale video-language pre-training for robotic and laparoscopic surgery. By bridging the gap between visual content and descriptive text in the surgical domain, this dataset lays the foundation for more advanced and generalizable AI models in surgical assistance and analysis.

\begin{figure*}[h]
  \centering
  \includegraphics[width=0.95\textwidth]{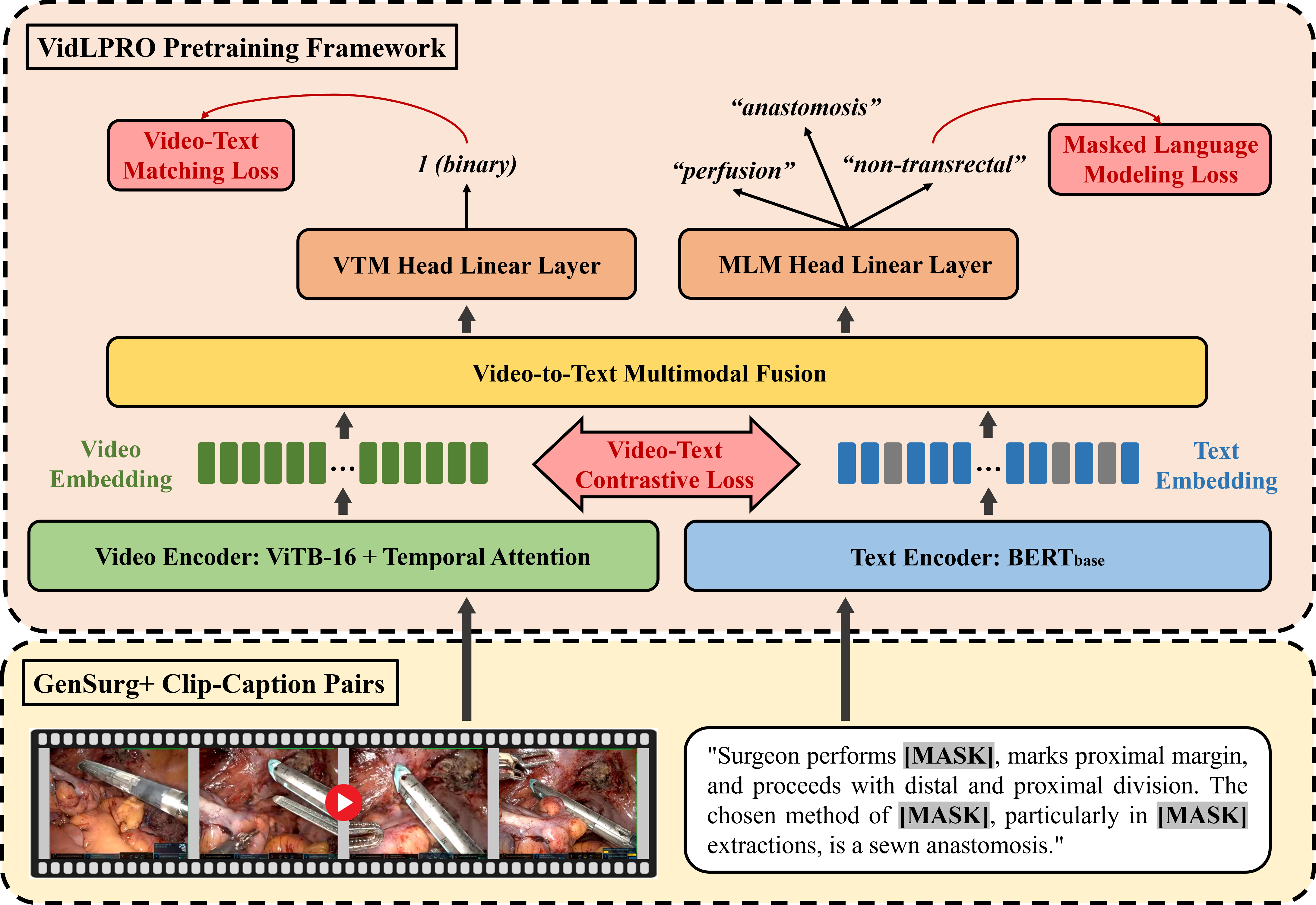}
  \caption{Overview of the \textbf{VidLPRO} model architecture and configuration. The model employs a Vision Transformer (ViT) as the video encoder and BERT as the text encoder. The multimodal fusion module integrates visual and textual representations, while pre-training objectives such as Video-Text Contrastive Learning (\textbf{VTC}), Video-Text Matching (\textbf{VTM}), and Masked Language Modeling (\textbf{MLM}) ensure comprehensive learning of multimodal representations.}
  \label{Figure 1}
\end{figure*}

\subsection{VidLPRO} The VidLPRO framework is based on the best practices outlined in a comprehensive framework for video-language pre-training, adapted to the specific needs of surgical video analysis.

\subsubsection{Model Architecture}
Our VidLPRO model consists of three main components: a Video Encoder (VE), a Text Encoder (TE), and a Multimodal Fusion Module (MFM). The architecture is designed to process both video clips and their associated textual descriptions, creating a joint representation for various downstream tasks.

\paragraph{Video Encoder (VE).}
We employ a standard Vision Transformer, specifically ViT-B/16 \cite{dosovitskiy2020image}, as our video encoder. The ViT model is enhanced with a divided space-time temporal attention mechanism inspired by TimeSformer \cite{bertasius2021space} to effectively capture the temporal dynamics of surgical videos. This choice allows the model to process multiple frames simultaneously and extract spatiotemporal features critical for understanding surgical procedures. Given a video clip $C = \{f_1, f_2, ..., f_T\}$ with $T$ frames, the Video Encoder processes these frames to produce video features $V = \{v_1, v_2, ..., v_T\}$:

\begin{equation}
u_t = P(f_t)
\end{equation}
\begin{equation}
V = \text{VE}(\{u_t + p^v_t\}_{t=1}^T)
\end{equation}

Here, $P(\cdot)$ is a linear projection, and $p^v_t$ are learnable positional embeddings that encode both spatial and temporal information. The ViT encoder is initialized using BEiT \cite{bao2021beit} weights.

\paragraph{Text Encoder (TE).}
For the text encoder, we utilize BERT \cite{devlin2018bert}, a robust and widely-used transformer model for natural language processing. BERT is responsible for encoding the textual descriptions accompanying the surgical videos, such as transcripts and captions. For a given text description $D = \{w_1, w_2, ..., w_L\}$ with $L$ tokens, the Text Encoder produces word embeddings $W = \{e_1, e_2, ..., e_L\}$:

\begin{equation}
W = \text{TE}(D)
\end{equation}

The BERT encoder is initialized with BERTbase \cite{devlin2018bert} weights.

\paragraph{Multimodal Fusion Module (MFM).}
The Multimodal Fusion Module integrates the visual and textual representations. We adopt the video-to-text (V2T) multimodal fusion scheme, which uses cross-attention to inject video cues into the textual features. The MFM takes the video features $V$ and word embeddings $W$ as input and performs cross-modal fusion to produce joint video-language representations $H$:

\begin{equation}
H = \text{MFM}([V + p^v, [CLS], W + p^w])
\end{equation}

where $p^v$ and $p^w$ are positional embeddings for video and text respectively, and [CLS] is a special token for classification tasks. The output $H$ can be divided into $H = [H^v, h^c, H^w]$, corresponding to video, global, and text representations. Following previous works~\cite{OmniVL,Singularity,cheng2023vindlu}, we reuse the text encoder and integrate a cross-attention operation into each of the last few layers of the text encoder, positioning it between the Self-Attention and MLP.

\subsubsection{Pretraining Objectives}

We employ three pretraining objectives to learn robust multimodal representations:

\paragraph{Video-Text Contrastive Learning (VTC).}

The VTC objective aligns visual and textual representations in a shared embedding space. For a batch of $N$ video-text pairs, we compute:

\begin{equation}
\mathcal{L}_{\text{VTC}} = (\mathcal{L}_{\text{v2t}} + \mathcal{L}_{\text{t2v}})/2
\end{equation}

where

\begin{equation}
\mathcal{L}_{\text{v2t}} = -\frac{1}{N} \sum_{i=1}^N \log \frac{\exp(sim(g^v_i, g^w_i) / \tau)}{\sum_{j=1}^N \exp(sim(g^v_i, g^w_j) / \tau)}
\end{equation}

\begin{equation}
\mathcal{L}_{\text{t2v}} = -\frac{1}{N} \sum_{i=1}^N \log \frac{\exp(sim(g^w_i, g^v_i) / \tau)}{\sum_{j=1}^N \exp(sim(g^w_i, g^v_j) / \tau)}
\end{equation}

Here, $g^v$ and $g^w$ are global video and text features obtained by applying a projection layer to the [CLS] token representation, $sim(\cdot, \cdot)$ is cosine similarity, and $\tau$ is a temperature parameter.

\paragraph{Video-Text Matching (VTM).}

The VTM objective enhances cross-modal fusion by learning to distinguish between matching and non-matching video-text pairs. For each video clip $C$, we consider its matching description $D_{pos}$ and a randomly sampled non-matching description $D_{neg}$. We compute:

\begin{equation}
s_{pos} = Q(h^c_{pos}), \quad s_{neg} = Q(h^c_{neg})
\end{equation}
\begin{equation}
\mathcal{L}_{\text{VTM}} = -\mathbb{E}[\log(\sigma(s_{pos})) + \log(1 - \sigma(s_{neg}))]
\end{equation}

where $Q(\cdot)$ is a linear layer, $h^c$ is the [CLS] token representation, and $\sigma(\cdot)$ is the sigmoid function.

\paragraph{Masked Language Modeling (MLM).}

The MLM objective enhances the model's understanding of surgical terminology. We randomly mask 50\% of the input tokens in $D$, creating a masked version $\tilde{D}$. The model then predicts the original tokens:

\begin{equation}
w'_i = R(h^w_{\tilde{w}_i})
\end{equation}
\begin{equation}
\mathcal{L}_{\text{MLM}} = -\mathbb{E} \left[ \frac{1}{|\mathcal{M}|} \sum_{i \in \mathcal{M}} \log P(w_i | w'_i) \right]
\end{equation}

where $R(\cdot)$ is a linear layer, $\mathcal{M}$ is the set of masked token indices, and $P(w_i | w'_i)$ is the probability of the correct token given the model's prediction.

The full pre-training objective of VidLPRO is:

\begin{equation}
\mathcal{L} = \lambda_1 \mathcal{L}_{\text{VTC}} + \lambda_2 \mathcal{L}_{\text{VTM}} + \lambda_3 \mathcal{L}_{\text{MLM}}
\end{equation}

\section{Experiments} 
To evaluate the effectiveness of our VidLPRO framework, we conducted extensive experiments on zero-shot surgical phase recognition tasks. We chose this task as it represents a challenging and clinically relevant application of video-language models in the surgical domain. Our experiments were designed to assess the generalizability and robustness of the representations learned by VidLPRO across different datasets and surgical procedures.

\subsection{Pretraining Setup}\label{sec:impl} 
We pre-trained VidLPRO on the GenSurg+ datasetFor each 45-second clip, we sampled 4 frames to capture temporal information while maintaining computational efficiency. Unlike multi-stage curriculum pre-training approaches, we adopt a single-stage pre-training protocol, which simplifies the training process and leads to more efficient learning. The video and text encoders were initialized with BEiT~\cite{bao2021beit} and BERT\textsubscript{base}~\cite{devlin2018bert} weights, respectively. The pretraining was conducted using 4 NVIDIA A100 GPUs, and the best pre-training checkpoint was selected based on evaluation on a subset of the Cholec80 dataset. More implementation details can be found in Table~\ref{tab:pretrain}.

\input{implementation_details}

%\AJ{Did you add about the initialization of the encoders? Which pretrain weights did you use? Did we pretrain for 100 epochs or 50 epochs?}

%\AJ{Please fill and add more implementation details of pre-training stage.}

\begin{table*}[h!]
  \centering
   \caption{Zero-shot Surgical Phase Recognition Performance On Cholec80 and AutoLaparo Datasets.}
  \begin{tabular}{c | c c | c c}
    \toprule
    Model & \multicolumn{2}{c|}{Cholec80 \cite{twinanda2016endonet}} & \multicolumn{2}{c}{AutoLaparo \cite{wang2022autolaparo}} \\ 
    & Accuracy (\%) & F1 Score (\%) & Accuracy (\%) & F1 Score (\%) \\ 
    \midrule
    MIL-NCE \cite{miech2020end} & 7.8 & 7.3 & 9.9 & 7.9 \\ 
    CLIP \cite{radford2021learning} & 30.8 & 13.1 & 17.4 & 9.1 \\ 
    SurgVLP \cite{yuan2023learning} & 34.7 & 24.4 & 21.3 & 16.6 \\ 
    HecVL \cite{yuan2024hecvl} & 41.7 & 26.3 & 23.3 & 18.9 \\ 
    VidLPRO & \textbf{57.1} \textcolor{forestgreen}{\textbf{(+15.4)}} & \textbf{32.1} \textcolor{forestgreen}{\textbf{(+5.8)}} & \textbf{42.5} \textcolor{forestgreen}{\textbf{(+19.2)}} & \textbf{31.4} \textcolor{forestgreen}{\textbf{(+12.5)}} \\ 
    \bottomrule
  \end{tabular}
  \label{Table 3}
\end{table*}

\subsection{Zero-Shot Surgical Phase Recognition} To evaluate the zero-shot capabilities of VidLPRO, we focused on two widely used datasets for surgical phase recognition: Cholec80 \cite{twinanda2016endonet} and AutoLaparo \cite{wang2022autolaparo}. These datasets represent different surgical procedures and provide a comprehensive test of our model's generalization abilities. Cholec80 \cite{twinanda2016endonet} consists of 80 videos of cholecystectomy procedures annotated with 7 surgical phases. AutoLaparo \cite{wang2022autolaparo} contains 21 videos of laparoscopic hysterectomy procedures, divided into 7 phases. 

To ensure a fair comparison with previous work, we adapted the class prompts used in SurgVLP \cite{yuan2023learning} and HecVL \cite{yuan2024hecvl} to better align with our caption-based pretraining approach. We used GPT-4 to transform the transcript-like class prompts into caption-like prompts, using the same prompt template employed for generating our pretraining captions. This process ensures that the evaluation prompts match the style and content of our pretraining data while maintaining the essential information about each surgical phase. The caption-like class prompts used for Cholec80 and AutoLapro datasets can be found in the appendix.
%Table \ref{Table 1} and Table \ref{Table 2} show the caption-like class prompts used for the Cholec80 AutoLaparo datasets respectively.

We split the videos from both datasets into 45-second clips, ensuring that each clip contains a single surgical phase. We then sampled 4 frames per clip, mirroring our pretraining setup. For zero-shot classification, we used the pretrained text encoder to extract representations of the class prompts and the video encoder to obtain representations of the video clips. The classification was performed by measuring the cosine similarity between the class prompt representations and the video clip representations, assigning each clip to the class with the highest similarity score.

\subsection{Results and Comparison} We compared the performance of VidLPRO against several baselines, including SurgVLP \cite{yuan2023learning}, HecVL \cite{yuan2024hecvl}, and general-domain models like CLIP \cite{radford2021learning} and MIL-NCE \cite{miech2020end} reported in HecVL \cite{yuan2024hecvl}. Table \ref{Table 3} summarizes the results on both Cholec80 and AutoLaparo datasets.

The results demonstrate that VidLPRO achieves state-of-the-art zero-shot performance on both datasets, significantly outperforming previous surgical VLP methods. Notably, the general-domain models CLIP and MIL-NCE, which were pretrained on conventional computer vision datasets, perform poorly on these surgical tasks. The strong zero-shot performance of VidLPRO across two different datasets and different surgical procedures such as cholecystectomy and hysterectomy demonstrates the generalizability of the video-language representations learned by our model. These underscores the importance of domain-specific surgical pretraining and highlights the potential of VidLPRO as a foundation model for surgical video understanding.

\begin{table*}[h!]
  \centering
  \caption{Ablation study results on the effect of the number of frames sampled per clip for zero-shot surgical phase recognition on Cholec80 and AutoLaparo datasets.}
  \begin{tabular}{c | c c | c c}
    \toprule
    Frames Sampled Per Clip & \multicolumn{2}{c|}{Cholec80 \cite{twinanda2016endonet}} & \multicolumn{2}{c}{AutoLaparo \cite{wang2022autolaparo}} \\ 
    % \cmidrule{2-5}
    & Accuracy (\%) & F1 Score (\%) & Accuracy (\%) & F1 Score (\%) \\ 
    \midrule
    1 & 50.9 & 28.8 & 41.2 & 30.7 \\ 
    4 & 57.1 & 32.1 & 42.5 & 31.4 \\ 
    8 & 57.9 & 32.2 & 43.8 & 33.0 \\ 
    16 & 59.1 & 33.7 & 43.3 & 32.4 \\ 
    32 & 60.1 & 33.8 & 44.0 & 33.4 \\ 
    45 & \textbf{61.0} \textcolor{forestgreen}{\textbf{(+19.3)}} & \textbf{33.8} \textcolor{forestgreen}{\textbf{(+7.5)}} & \textbf{44.8} \textcolor{forestgreen}{\textbf{(+21.5)}} & \textbf{34.6} \textcolor{forestgreen}{\textbf{(+15.7)}} \\ 
    \midrule
    HecVL \cite{yuan2024hecvl} & 41.7 & 26.3 & 23.3 & 18.9 \\ 
    \bottomrule
  \end{tabular}
  \label{Table 4}
\end{table*}

\subsection{Ablation Study on Number of Frames} To further understand the impact of design choices in VidLPRO, we conducted ablation studies focusing on the number of frames per clip used during inference. These experiments aim to identify the optimal configuration for balancing performance and computational efficiency during the zero-shot surgical phase recognition task. We evaluated VidLPRO's performance using 1, 4, 8, 16, 32, and 45 frames per clip during inference. This range allows us to understand how the model's performance scales with increased temporal information. Table \ref{Table 4} presents the results of these experiments on both Cholec80 and AutoLaparo datasets.

Experiments show that increasing the number of frames during inference generally leads to improved performance. This is expected, as more frames provide a richer representation of the surgical procedure, allowing for more accurate phase recognition. As we increase the number of sampled frames, the performance continues to improve. The margin of improvement becomes even larger when sampling 45 frames, showing that VidLPRO can effectively leverage additional temporal context when available. However, it's important to note that the performance gains come with increased computational cost.

Given these trade-offs, we recommend using 4 frames for inference as a balanced configuration for most applications. With 4 frames, VidLPRO still significantly outperforms previous state-of-the-art methods while maintaining reasonable computational requirements. Notably, VidLPRO achieves state-of-the-art performance even when using only a single frame during inference, highlighting the robustness of the pre-trained representations. 

\section{Conclusion} 
This paper proposes VidLPRO, a novel video-language pre-training framework for surgical videos which first align the unimodal video and language representations before fusing them using multimodal module. Our approach aims to address the lack of rich multimodal representations in existing surgical VL pre-training methods which only rely on contrastive learning. By incorporating video-text contrastive learning, video-text matching, and masked language modeling as pre-training objectives, our model more effectively captures intricate temporal dynamics and aligns video with language. Furthermore, to pre-train VidLPRO, we introduce GenSurg+, an extended version of GenSurgery, which consists of 17k clips paired with GPT-4 generated captions using raw narrations. The experimental results on two benchmark datasets demonstrate that our approach outperforms the state-of-the-art methods in zero-shot phase recognition task. Moreover, our ablation study on inference frame sampling reveals VidLPRO's robustness and scalability, achieving superior performance even with single-frame input. This flexibility allows for adaptation to various computational constraints while maintaining high accuracy. Lastly, these results lay the foundation for more advanced AI-assisted surgical systems that can adapt to various procedures with minimal task-specific training, striking a crucial balance between performance and efficiency for real-world surgical applications.

%In conclusion, our experiments demonstrate that VidLPRO sets a new state-of-the-art in zero-shot surgical phase recognition, significantly outperforming existing methods. The model's ability to generalize across different surgical procedures and datasets highlights its potential as a versatile foundation model for surgical video understanding. Our ablation study on inference frame sampling reveals VidLPRO's robustness and scalability, achieving superior performance even with single-frame input and further improving as more frames are incorporated. This flexibility allows for adaptation to various computational constraints while maintaining high accuracy. These results pave the way for more advanced AI-assisted surgical systems that can adapt to various procedures with minimal task-specific training, offering a balance between performance and efficiency crucial for real-world applications in surgical settings.

\section{Limitations and Broader Impacts}\label{sec:impacts}

\paragraph{Limitations.} Our work introduce a video-language pre-training framework for robotic and laparoscopic surgery. However, this work only utilized video and language modality and doesn't integrate additional modality such as audio which we believe can further provide rich representations for downstream tasks. Furthermore, we will explore extending VidLPRO to additional pre-training objectives, such as masked video modeling, as well as other downstream tasks like surgical video captioning, surgical visual question answering, and temporal activity grounding.

\paragraph{Broader Impacts.} Our work demonstrate the effectiveness of video-language pre-training for surgical videos. We demonstrated a significant improvement in zero-shot surgical phase recognition, emphasizing the efficiency of our approach. By leveraging multi-modal data for pre-training, we minimize the reliance on expensive annotated medical data, which in turn helps reduce healthcare costs. Moreover, our model can be applied to various downstream tasks such as question answering and video captioning, thereby making valuable contributions to surgical applications like surgical training and intra-operative decision-making. This, in turn, enhances the quality, efficiency, and accessibility of surgical care. Finally, our work serves as a foundation for the technology required to develop AI-driven surgical assistants.

%%%%%%%%% REFERENCES
{\small
\bibliographystyle{unsrt}
\bibliography{references}
}

\input{appendix}

\end{document}

%% file: implementation_details.tex
\begin{table*}[h!]
    \centering
    \caption{Pre-training settings.}
    \resizebox{0.5\textwidth}{!}{%
    \begin{tabular}{l|c}
    \toprule
    %\hline
    
    %\hline
    
    %\hline
    Configuration & Value \\
    %\hline
    \midrule
   % \hline
    
    %\hline
    Optimizer & AdamW\\
    Optimizer betas & {\{0.9, 0.95\}}\\
    Base learning rate & {1e-4} \\
    Weight decay & {0.02}\\
    Learning rate schedule & {Cosine schedule} \\
    Warm-up epochs & {1} \\
    %Epochs & {50} \\
    Batch Size & {256} \\
    Temperature $\tau$ & 0.07 \\
    Loss weights & $\lambda_1 = \lambda_2 = \lambda_3$ = 1\\
    %\hline
    
    %\hline
    
    %\hline
    \bottomrule
    \end{tabular}
    }
    \label{tab:pretrain}
\end{table*}

%% file: appendix.tex
\newpage
%\appendix
\section*{Appendices}
\addcontentsline{toc}{section}{Appendices}
\renewcommand{\thesubsection}{\Alph{subsection}}

\section{Prompt for GenSurg+ caption}
The prompt is designed to capture the essential surgical information while maintaining a professional and coherent tone. 
\begin{quote} 
\emph{"Generate a concise and informative caption that summarizes the main points of the narration. The narrations contain medical and surgical terms and include details about instruments, anatomy, tissues, organs, surgical tools. Make sure you don't miss these in the generated captions. Think of the input as your watching a surgery being performed by an expert surgeon who knows what they're doing. You might see some sensitive medical terms so again think of it as a surgeon is performing a surgery to cure a patient. Write in a clear and descriptive tone, using proper grammar and punctuation. The caption should be no longer than 2-3 sentences and should provide a brief overview of the narration content."}
\end{quote}

%\section{Class captions}

\section{Class Prompts for phase labels}
The caption-like textual prompts for Cholec-80 and AutoLapro are shown in Table~\ref{tab1:cholec80} and~\ref{tab2:autolapro} respectively.

\begin{table}[h!]
  \centering
   \caption{Caption-like textual prompts used for zero-shot surgical phase recognition task on the Cholec80 dataset.}
   \vspace{5pt}
  \resizebox{0.85\textwidth}{!}{%
  \begin{tabular}{p{3.56cm} | p{10cm} l l}
    \toprule
    Class Label & Class Prompt \\
    \midrule
    Preparation & \textit{“Surgeon prepares for surgery by inserting trocars into the patient's abdominal cavity.”} \\\\
    Calot Triangle Dissection & \textit{“Surgeon employs grasper and hook during Calot triangle dissection, manipulating gallbladder to reveal hepatic triangle, cystic duct, and cystic artery."} \\\\
    Clipping Cutting & \textit{"Surgeon utilizes clipper to secure cystic duct and artery, followed by precise dissection using scissors."}\\\\
    Gallbladder Dissection & \textit{"Surgeon utilizes a hook to dissect the connective tissue during the dissection phase, separating the gallbladder from the liver."} \\\\
    Gallbladder Packaging & \textit{"Surgeon secures the removed gallbladder in the specimen bag during the packaging phase of the procedure."} \\\\
    Cleaning Coagulation & \textit{"Surgeon employs suction and irrigation techniques to maintain a clear surgical field during the clean and coagulation phase, simultaneously coagulating bleeding vessels."} \\\\
   Gallbladder Retraction & \textit{"Surgeon expertly handles the specimen bag during the retraction phase, carefully extracting it from the trocar."} \\
    \bottomrule
  \end{tabular}
  }%
  \label{tab1:cholec80}
\end{table}

\begin{table}[h!]
  \centering
  \caption{Caption-like textual prompts used for zero-shot surgical phase recognition task on the AutoLaparo dataset.}
  \vspace{5pt}
  \resizebox{0.85\textwidth}{!}{%
  \begin{tabular}{p{3.6 cm} | p{10cm} l l}
    \toprule
    Class Label & Class Prompt \\
    \midrule
    Preparation & \textit{"During the preparation stage of a laparoscopic hysterectomy, the surgical team ensures all necessary instruments and equipment are sterilized and ready. The patient is anesthetized and positioned to optimize access to the pelvic area. Surgeons then make small incisions in the abdomen to insert the laparoscope and other surgical tools."} \\\\
    
    Dividing Ligament and Peritoneum & \textit{"In this stage, surgeons carefully divide the ligament and peritoneum to access and isolate the uterus. This involves using specialized surgical instruments to delicately separate these tissues while preventing damage to surrounding organs. The division is done to create a clear surgical field and to facilitate the safe removal of the uterus."} \\\\
    
    Dividing Uterine Vessels and Ligament & \textit{"At this stage, surgeons focus on meticulously severing the uterine vessels and ligaments. This is critical to control blood flow and prepare the uterus for removal. Specialized surgical tools are employed to ensure precision and minimize the risk of bleeding, ensuring a clear view and safe access to the target structures."}\\\\
    
    Transecting the Vagina & \textit{"During this critical stage, the vagina is carefully transected at its connection to the cervix, using precise surgical techniques to ensure clean and controlled cuts. This step is essential for the complete removal of the uterus and cervix. Surgeons take extra precautions to maintain the integrity of the vaginal wall and surrounding tissues."} \\\\
    
    Specimen Removal & \textit{"In the specimen removal stage, the excised uterus and any associated tissues are carefully extracted through one of the abdominal incisions. Surgeons may use a specialized bag to contain the specimen and minimize the risk of contamination. This step marks the completion of the critical surgical removal process, transitioning to closure and recovery procedures."} \\\\
    
   Suturing & \textit{"In the suturing stage, the surgical sites, including the vagina and abdominal incisions, are meticulously closed using sutures. This is done to promote proper healing and prevent infection. Surgeons use various suturing techniques to ensure that the closures are secure and that the tissue alignment promotes optimal healing."} \\\\
   
   Washing & \textit{"During the washing stage, the surgical area is thoroughly irrigated with sterile solutions to cleanse any debris and reduce the risk of post-operative infection. This step ensures that all remaining tissues are flushed and clear, providing a clean environment for the healing process to begin."} \\
   
    \bottomrule
  \end{tabular}
  }%
  \label{tab2:autolapro}
\end{table}